\documentclass[letterpaper]{article} 
\usepackage{aaai25}  
\usepackage{times}  
\usepackage{helvet}  
\usepackage{courier}  
\usepackage[hyphens]{url}  
\usepackage{graphicx} 
\usepackage{natbib}  
\usepackage{caption} 
\usepackage{algorithm}
\usepackage{algorithmic}

\usepackage{newfloat}
\usepackage{listings}
\DeclareCaptionStyle{ruled}{labelfont=normalfont,labelsep=colon,strut=off} 
\floatstyle{ruled}
\newfloat{listing}{tb}{lst}{}
\floatname{listing}{Listing}
\title{Towards Automated Scoping of AI for Social Good Projects}
\author{
    Jacob Emmerson\textsuperscript{\rm 1}, Rayid Ghani\textsuperscript{\rm 2}, Zheyuan Ryan Shi\textsuperscript{\rm 1}
}
\affiliations{
    \textsuperscript{\rm 1}University of Pittsburgh, \textsuperscript{\rm 2}Carnegie Mellon University\\

    \{jte27, ryanshi\}@pitt.edu, rayid@cmu.edu
}

\usepackage{amsmath}
\usepackage{csquotes}
\usepackage{colortbl}
\usepackage{soul}
\usepackage{tcolorbox}
\definecolor{wtbl}{gray}{1}
\definecolor{gtbl}{gray}{0.9}

\begin{document}

\maketitle

\begin{abstract}
Artificial Intelligence for Social Good (AI4SG) is an emerging effort that aims to address complex societal challenges with the powerful capabilities of AI systems. These challenges range from local issues with transit networks to global wildlife preservation. However, regardless of scale, a critical bottleneck for many AI4SG initiatives is the laborious process of problem scoping---a complex and resource-intensive task---due to a scarcity of professionals with both technical and domain expertise. Given the remarkable applications of large language models (LLM), we propose a Problem Scoping Agent (PSA) that uses an LLM to generate comprehensive project proposals grounded in scientific literature and real-world knowledge. We demonstrate that our PSA framework generates proposals comparable to those written by experts through a blind review and AI evaluations. Finally, we document the challenges of real-world problem scoping and note several areas for future work. 

\end{abstract}

%

\section{Introduction}

AI for Social Good (AI4SG) has received considerable interest from the academia, industry, and government sectors~\cite{Floridi2020HowTD,shi2020artificialintelligencesocialgood}. This initiative uses AI methods to address societal issues, such as healthcare inequality~\cite{AIinHEALTHCARE,OnAT}, disparities in public transit \cite{Liu2020APA,Park2022MultiobjectiveAT}, and climate change \cite{Xu2020DualMandatePB,Sisodia2023AITI,10.5555/2832581.2832611}.

Over the past decades, the effort, resources, and goodwill poured into this initiative have achieved great success, and also revealed many key bottlenecks. A key bottleneck is problem scoping, where the problem owners and stakeholders identify a pain point and how a particular AI method might solve it. Problem scoping can be a time-consuming and laborious process. It requires understanding the limitations of different technological methods and the constraints of the social issue at hand.
Many public sector organizations lack the technical expertise to understand the former, while technologists are often negligent about the latter. When public sector organizations want to devote their already limited budget to technology projects, they have a hard time finding the relevant technical expertise, and end up with either missed opportunities or ill-scoped projects that cost them more down the road. Arguably one of the biggest successes of AI4SG initiatives is the emergence of researchers and practitioners who can bridge this gap. Yet they are too scarce in numbers compared to the abundant societal challenges and public sector organizations. 
\textit{If one could scale up this resource, even if at the cost of, say, only 60\% as effective as experienced scoping professionals, one could greatly facilitate the promise and impact of AI4SG.}

We explore whether generative models can alleviate the resource requirements for problem scoping to promote future AI4SG initiatives.
Large language models (LLMs) have been applied as mathematical provers \cite{Xin2024DeepSeekProverAT,Han2021ProofAC,Jiang2022ThorWH,Jiang2022DraftSA}, research tools \cite{lu2024aiscientistfullyautomated,pu2024ideasynthiterativeresearchidea,baek2024researchagentiterativeresearchidea,si2024llmsgeneratenovelresearch}, and code generators \cite{ifargan2024autonomousllmdrivenresearchdata,gu2024bladebenchmarkinglanguagemodel,hu2024infiagentdabenchevaluatingagentsdata,lu2024aiscientistfullyautomated,li2024mlrcopilotautonomousmachinelearning}. Their inherent knowledge and ability to mimic human-like behaviors suggests they could automate or accelerate structured, information-intensive tasks. Problem scoping is one such task; it can be broken down into several general and repetitive steps requiring new information. 
Furthermore, unlike previous generations of AI methods, LLMs have been widely used by professionals who do not have technology expertise, including many in the public sector. As a result, it is much more natural for them to use LLMs to generate AI project ideas within their organizations.

Admittedly, this affinity to LLMs is a double-edged sword. It is well-known that LLMs often fail to properly ground knowledge in a given domain, and could even generate false or inappropriate content. As we demonstrate later in the paper, using LLMs for scoping in a naive fashion would not be ideal. 
In this work, we use LLMs grounded by various information sources and queries to automate the scoping process. 
We make the following contributions. 
(1) We introduce the AI4SG problem scoping task, which has high real-world relevance and is also distinct to existing LLM tasks. (2) We propose the Problem Scoping Agent (PSA), an LLM pipeline enhanced by retrieval-augmented generation for coming up with AI4SG problem scopes.
(3) We run experiments to compare our PSA with vanilla LLM methods and expert human professionals, and 
show that PSA achieves better scoping outcome than vanilla LLM methods and not too far from human experts in many cases.

\begin{figure*}[t!]
    \centering
    \includegraphics[width=\linewidth]{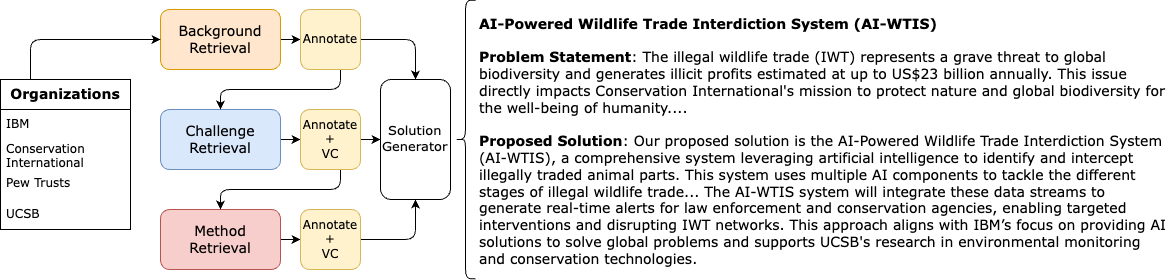}
    \caption{Problem Scoping Agent Overview}
    \label{fig:psa_overview}
\end{figure*}

\section{Related Work}
\subsubsection{Idea Generation with AI Agents.}
Several prior studies evaluated the potential to accelerate research using generative models. Frameworks proposed in previous work generally take an iterative approach, where the current ideas are updated according to an outlined criteria and provided knowledge base. This knowledge base may be actively queried through an API or constructed before generating ideas. Iterative refinement \cite{lu2024aiscientistfullyautomated,baek2024researchagentiterativeresearchidea,pu2024ideasynthiterativeresearchidea,si2024llmsgeneratenovelresearch}, multimodule retrieval with feedback \cite{yang2024largelanguagemodelsautomatedmultimodule}, and facet recombination \cite{radensky2024scideatorhumanllmscientificidea} have shown success as fully automated systems, while other works focus on a collaborative procedure that improves brainstorming \cite{lee2024conversationalagentscatalystscritical}.  Similarly, several studies have tested whether generative models can operationalize or implement research projects by generating code and automating data analysis \cite{ifargan2024autonomousllmdrivenresearchdata,gu2024bladebenchmarkinglanguagemodel,hu2024infiagentdabenchevaluatingagentsdata,lu2024aiscientistfullyautomated,li2024mlrcopilotautonomousmachinelearning}. Distinct from these works, idea generation in a social-good context carries higher stakes and generally emphasizes less novel solutions in exchange for tractability and effectiveness. Most similar to our work is \cite{zhao2024foundationmodelbasedmultiagentaccelerateai} where the authors outline the idea of foundation models as catalysts for collaborative social change. Differing from their work, we propose a fully-automated system and emphasize an exploration of problem scoping under a multiple contexts rather than resource allocation alone. 

\subsubsection{LLMs as Automatic Evaluators.}
Several studies have additionally explored the use LLMs as  evaluators and reviewers. While LLMs demonstrate a non-trivial performance on pair-wise comparisons \cite{si2024llmsgeneratenovelresearch}, it is unclear whether such models can empirically evaluate solutions. Some work suggests that LLM achieves a higher agreement with human reviewers than reviewers themselves \cite{lu2024aiscientistfullyautomated}; however, other studies find that the agreement between human reviewers and LLMs is less than random \cite{si2024llmsgeneratenovelresearch,stureborg2024largelanguagemodelsinconsistent}. Furthermore, LLMs demonstrate a preference for generations with a lower perplexity, typically their own generations, over human-written text \cite{panickssery2024llmevaluatorsrecognizefavor}. Additional studies report that LLMs are sensitive to the prompts used during evaluation \cite{stureborg2024largelanguagemodelsinconsistent}, indicating that LLMs are not yet capable of performing consistent human-quality evaluations.

\begin{figure*}[t!]
    \centering
    \includegraphics[width=\linewidth]{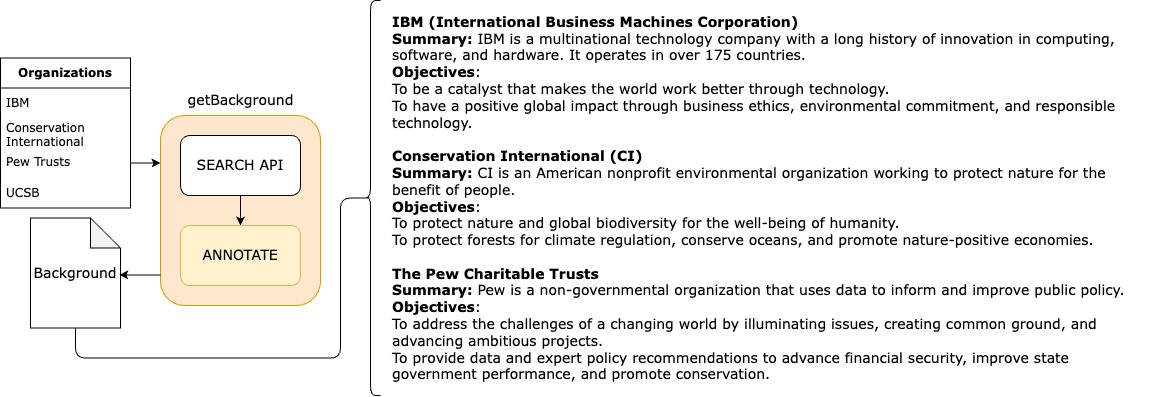}
    \caption{Background Retriever}
    \label{fig:background}
\end{figure*}

\section{Problem Scoping Agent}

While there exists a plethora of literature on a variety of AI methods, it can be time consuming to implement and evaluate existing solutions on problems with constraints beyond those encountered in traditional benchmarks. Furthermore, though the limitations of some methods may be documented, connecting those limitations with problem constraints can be a non-trivial process. To this end, we leverage techniques from AI idea generation and prompt-engineering methods to evaluate the extent in which LLMs can:
\begin{enumerate}
    \item \textbf{Identify} tractable, real-world social good problems faced by public sector organizations.
    \item \textbf{Propose} relevant methods from AI/ML that are appropriate for the problem.
    \item \textbf{Integrate} a solution that is properly contextualized on the problem.
\end{enumerate}

Our approach to automated problem scoping is unique as we seek to integrate the theoretical methods mentioned in research with real-world problems, problems that can be more complex due to unforeseen constraints. Thus, we introduce a Problem Scoping Agent (PSA), an automated approach to identifying and proposing solutions to challenges that technologically-sparse organizations may face. Technologically-sparse organizations may lack an in-house analytics and engineering team, making the integration of AI methods risky and time-consuming. 

As shown in Fig.~\ref{fig:psa_overview}, this task is decomposed into series of steps where the agent determines the organization's background, challenges they may face, and appropriate methods for an identified problem. This ensures that the generation of a solution proposal can be guided by any nuanced details that may otherwise go unaccounted for in a one-shot setting. Intuitively, this is similar to chain-of-thought: We build a solution from the ground up, passing forward only the relevant information at each step to guide our queries. The `Memphis Fire Department' is used in each figure as a running example, though the generated outputs are truncated for clarity. The exact prompts used can be found in Appendix \ref{appendix:prompts}. Subsequently, we describe each component of PSA.

\subsubsection{Annotator.} The PSA relies on continuous API calls to retrieve relevant information; however, the retrieved documents can be very noisy. This makes any important details within the documents harder to extract when given a large corpora of text; to overcome this, we employ an LLM to annotate and summarize any retrieved articles. While significantly helpful for cleaning information pulled from web pages, we additionally annotate any academic papers for an easier comparison between problems and their potential solutions by extracting the target problem, methods, field, and outcome of the work.

\subsubsection{Background.} 

Given a list of organization names $O = \{o_1\dots o_N\}$, we identify the relevant background and goals, $B$, of each organization. Using a function $B = getBackground(O)$, a we search for the historical objectives of each $o_i \in O$ as well as any relevant information using the Google Custom Search API. 

Concretely, the LLM retrieves $3$ web pages for the $N$ organizations involved in the proposal. Each page is summarized using our annotator into a concise statement $B$ that can then be used to inform our queries when searching for relevant challenges. In Figure \ref{fig:background}, the agent searches for any background information on $O = \{\text{IBM, Conservation International, Pew Trusts, UCSB}\}$ and returns a background, $B$, summarizing the objectives and previous initiatives of the Memphis Fire Department. Some information in Figure \ref{fig:background} is omitted for clarity.

\subsubsection{Challenges.} 

Identifying the challenges relevant to an organization is fundamental to this project, particularly challenges that are reasonably tractable given the resources available to $O$. 

\begin{figure*}[t!]
    \centering
    \includegraphics[width=.9\linewidth]{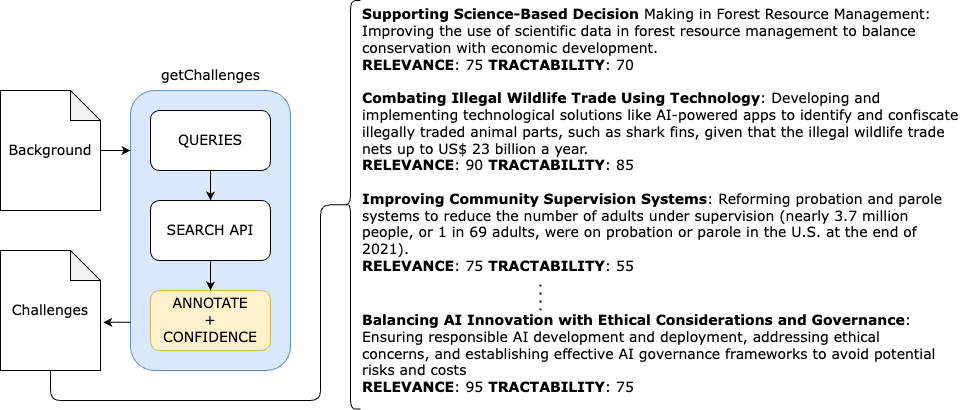}
    \caption{Challenge Retriever}
    \label{fig:challenges}
\end{figure*}

Due to their tendency for hallucinations and overall lack of diversity during generation \cite{si2024llmsgeneratenovelresearch}, LLMs struggle to determine what problems remain active. This particularly true for problems that are domain-specific or local to $O$. 
To increase the diversity of identifiable problems and ensure they have some level of factuality, we query Google Search API for 5 potential challenges using LLM-generated challenges. The retrieved challenges $C = getChallenges(B)$ should be relevant to the organization's background, though they need not be previously worked on by the organization. They must additionally contain sufficient detail to be motivated and integrated with the proposed solution as shown in Figure \ref{fig:challenges}. For each $c \in C$, we estimate the model's confidence and average the estimate of $c$'s relevance and tractability to the organization. This estimate is obtained using verbalized confidence \cite{yang2024verbalizedconfidencescoresllms}, although any sampling-based or black-box uncertainty quantification methods could also be used. 

\subsubsection{Methods.} Guided by a sampled $c \in C$, a query is made to Semantic Scholar's API to search for AI methods specifically relevant to $c$. The challenge is sampled from a distribution with probabilities proportional to the average confidence estimates of each $c$, where the estimated confidences are turned into a distribution using the softmax function.
In our work, we find that LLMs demonstrate a high-level understanding of AI, machine learning, and statistics, but they lack the human-level considerations---bias, resource constraints, etc---when selecting such methods as a solution. While not empirically demonstrated in this study, this challenge is adjacent to the known limitations of LLMs on code generation \cite{chen2021evaluatinglargelanguagemodels,jimenez2024swebenchlanguagemodelsresolve} and self-interpretation \cite{sherburn2024languagemodelsexplainclassification, 10.1145/3442188.3445922}. They lack an understanding of the true ``function" or implications of different methods and struggle to disentangle them from the contexts in which they are used.

\begin{figure*}[t]
    \centering
    \includegraphics[width=\linewidth]{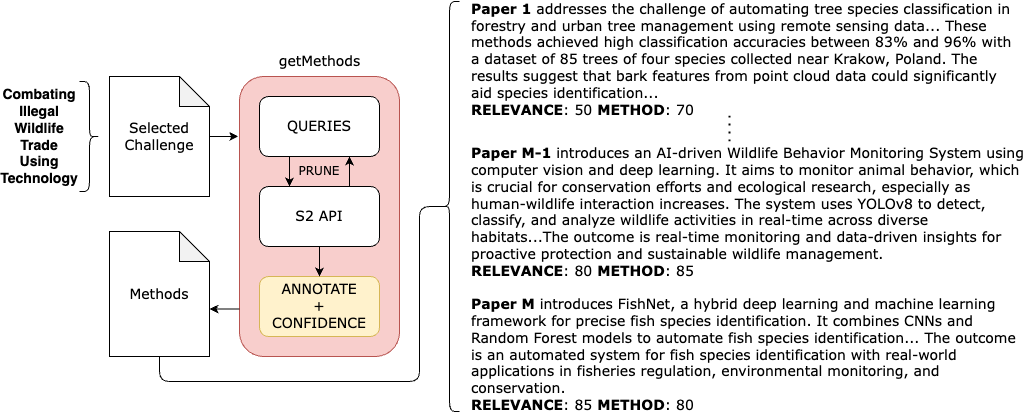}
    \caption{Method Retriever}
    \label{fig:methods}
\end{figure*}

To provide the model with context of how different methods are used, we retrieve a cohort of literature $M$ that contextualizes relevant AI methods using LLM-generated queries to Semantic Scholar. Concretely, the LLM generates 5 search queries for conference papers or journal articles that are related to $c$ and retrieves up to $|M| = 10$ papers. Since some challenges may not contain sufficient prior work in AI, resulting in no previously proposed solutions, we search for solutions to the fundamental problem of the challenge---resource allocation, classification, and network optimization are common examples of fundamental problems. Despite this, there may still be a lack of literature for a generated query. If zero papers are returned, the query is pruned and used again; as the queries are pruned, they typically become more general. For example, Figure \ref{fig:methods} uses a query ``statistical techniques for environmental impact analysis and mitigation" which may have no returnable papers, but there is likely to be work on ``statistical techniques for environmental impact". After retrieving a non-zero number of papers, these are annotated and prompted for a confidence score of each paper's relevance and applicability to the provided challenge. During the final generation, only 5 papers with the highest confidence are provided to avoid overwhelming the models with unrelated details.

This process encourages a divergent-convergent reasoning approach over different AI methods; this is a technique commonly used to overcome a cognitive bias known as ``functional-fixedness" that prevents humans from applying tools or techniques in domains outside of which they are observed. By taking specific examples from prior work and generalizing them to a broader context, they can be more easily aligned with different problem scenarios. Previously, this has been applied as a prompting technique for the related task of creative problem solving \cite{tian2024macgyverlargelanguagemodels}.

\subsubsection{Solution Generation.} While the search queries are guided by only the most relevant information at each step, the final proposal generation leverages all of the relevant annotated information to ensure that any relevant statistics, techniques, or model names are present. Aligning the limitations of different methods with problem constraints is the primary task of this module since out-of-domain work may be applied on domain-specific problems. Succinctly, the proposed solution should focus on the constraints identified in $c$, limitations of each $m \in M$, and frame the proposal relevant to $B$. 

\subsubsection{Verbalized Confidence.}

Verbalized confidence \cite{yang2024verbalizedconfidencescoresllms,xiong2024llmsexpressuncertaintyempirical} is a straightforward approach to quantifying the confidence an LLM has in its output by simply prompting for an estimate. Despite the tendency for models to be overconfident in their estimates, we prompt for the confidence of the \emph{relevance} for each $c \in C$ and each $m \in M$ to help make the model's decision making process more transparent. Additionally, we prompt for the \emph{tractability} confidence for the former and \emph{applicability} confidence for the latter. We expect the chosen method to be relevant and applicable on a selected challenge, a challenge which should be relevant and tractable given the background of an organization.

\section{Experiments}
\subsection{Setup}
\subsubsection{Dataset.}

For comparison against projects thought up by real people, we use $|X| = 21$ project summaries from Data Science for Social Good (DSSG)\footnote{\url{https://www.dssgfellowship.org/}} for our evaluation. 
DSSG is a well-recognized program that pairs students with public sector organizations to work on data science projects. The program has been successfully run for over 10 years. Each project at DSSG undergoes a rigorous scoping process guided by an expert AI4SG researcher, which ensures the high quality of the output.
For each project summary, we extract the involved organizations and summarize the project description according to our format in Appendix \ref{appendix:formats}.

\subsubsection{Models.}

We generate proposals using three different LLMs as the core component of the PSA: GPT-4o, Gemini-2.0-Flash, and DeepSeek-V3. Each generation is sampled with a temperature $\tau = 0.9$. GPT-4o is additionally used to rewrite the original proposals to ensure consistent formatting and a fair evaluation. The prompt for rewriting is the same used for combining the final proposal which can be found in Appendix~\ref{appendix:prompts}. 

\subsubsection{Evaluation Metrics.}

Following the current work on idea generation, we run a blind evaluation using a set of predefined metrics. We used a modified version of the criteria outlined in \cite{si2024llmsgeneratenovelresearch}. The original criteria---Novelty, Excitement, Feasibility, and Expected Effectiveness---evaluated AI-generated novel research ideas. For this work, we focus on finding potentially novel, real-world \emph{applications} of prior work. To this end, we defined a modified variant of their criteria on a Likert scale (1-5):
\begin{itemize}
    \item \textbf{Appropriateness:} How related is the proposal to the original problem domain? Is the identified problem related to the organization's goals? Does this problem necessitate the use of AI methods? Are the proposed methods appropriate for the identified problem?  
    \item \textbf{Thoroughness:} The concrete-ness of the proposed solution. Are any resource requirements specified? Is the solution justified, and does the justification make technical sense? Are the methods tied in with the identified problem (e.g. for resource allocation, what resources are being allocated)?
    \item \textbf{Feasibility:} Is the identified problem reasonably tractable? How much progress can be made within the first few weeks? How will success be measured throughout the project?
    \item \textbf{Expected Effectiveness:} The impact on the identified problem given a successful implementation of the proposed solution. What will change, and how much will it change? Is this project sustainable? Will this solution continue to be useful for the target demographics within the next decade?
\end{itemize}

To ensure fairness, each statement is evaluated independently with no notion of the method used for its creation or what future proposals may look like. The order in which the statements are shown is permuted to help prevent an ordering bias.

We also evaluate each proposal using AI for comparison against a blind reviewer on our modified criteria.
Considering recent criticism on the use of generative AI for automated evaluations due to their inconsistency and low correlation with human reviewers \cite{si2024llmsgeneratenovelresearch,stureborg2024largelanguagemodelsinconsistent}, each response is sampled 3 times to gauge the uncertainty of the evaluations. 
The exact prompts used for evaluation can be found in Appendix \ref{appendix:formats}
Specifically, we use GPT-4o-2024-08-06 and Claude-3.5-Sonnet-20240620 for evaluation with the same parameters used during proposal generation. Claude is specifically chosen for unbiased evaluations given its success in \cite{si2024llmsgeneratenovelresearch} and Claude not being used for proposal generation~\cite{panickssery2024llmevaluatorsrecognizefavor}. 
However, the outcome of LLM-based evaluation is not satisfactory due to a low variance in the scores between all proposals and and low correlation with Human evaluations as reported in Table \ref{tab:correlations}. Instead, we include the LLM-based evaluations in Appendix \ref{appendix:ai_evaluations}.

\subsection{Results}

\begin{table}[hbt]
    \centering
    \begin{tabular}{l c c}
    \hline
    Model & $\rho_{\text{ Human-AI}}$ & $\sigma^2_{\text{all scores}}$\\
    \hline\hline
    Human & - & 0.4923 \\
    DeepSeek-R1 & 0.2134 & 0.0172 \\
    Claude-3.5-Sonnet & -0.0465 & 0.0518 \\
    GPT-4o & 0.0248 & 0.0361 \\
    Ensemble & 0.0233 & 0.0244 \\
    \hline
    \end{tabular}
    \caption{Correlation coefficients between averaged AI and human evaluations and variance of AI evaluations.}
    \label{tab:correlations}
\end{table}

We evaluated proposals generated with DeepSeek-V3, GPT-4o-2024-08-06, and Gemini-2.0-Flash in addition to proposals generated by their respective scoping agents DS-PSA, GPT-PSA, and G-PSA. Our primary interest is whether LLMs and their agents can generate proposals comparable to Human designed proposals. 
To compare the score distributions between our approaches, we report the $p$-values from a paired sample $t$-test for each individual metric and a multivariate Hotelling's $T^2$ test when comparing the center (average) of the distributions. The latter accounts for the covariance within each metric, information that would otherwise be unaccounted for through a univariate test on the average. 

\begin{table*}[tb]
    \centering
    \rowcolors{1}{wtbl}{gtbl}
    \begin{tabular}{l c c c c c}
        \hline
        Model & Appropriateness & Feasibility & Thoroughness & Expected Effectiveness & Average* \\
        \hline\hline
        Original & 4.0952 \small$\pm$ 0.56 & 3.8571 \small$\pm$ 0.50 & 3.8095 \small$\pm$ 0.54 & 3.6667 \small$\pm$ 0.89 & 3.8571 \small$\pm$ 0.35 \\ 
        \hline
        DeepSeek-V3 & -0.7143 \small$\pm$ 0.49 & -0.7143 \small$\pm$ 0.68 & -0.4762 \small$\pm$ 1.11 & -0.5714 \small$\pm$ 1.01 & -0.6190 \small$\pm$ 0.33 \\ 
        & 0.0002 & 0.0009 & 0.0565 & 0.0192 & 0.0017 \\
        Gemini-2.0 & \textbf{0.5714 \small$\pm$ 0.72} & 0.7619 \small$\pm$ 0.56 & 0.2381 \small$\pm$ 0.85 & \textbf{0.7619 \small$\pm$ 0.94} & \textbf{0.5833 \small$\pm$ 0.37} \\ 
        & 0.0069 & 0.0002 & 0.2612 & 0.0022 & 0.0022 \\
        GPT-4o & -0.3810 \small$\pm$ 1.38 & -0.2857 \small$\pm$ 1.35 & -0.2381 \small$\pm$ 1.51 & -0.0476 \small$\pm$ 1.66 & -0.2381 \small$\pm$ 1.05 \\ 
        & 0.1623 & 0.284 & 0.3972 & 0.8705 & 0.5171 \\
        \hline
        DS-PSA & -0.0476 \small$\pm$ 0.90 & -0.1905 \small$\pm$ 0.92 & 0.1429 \small$\pm$ 0.60 & -0.0476 \small$\pm$ 1.09 & -0.0357 \small$\pm$ 0.34 \\ 
        & 0.8249 & 0.3841 & 0.4187 & 0.8406 & 0.6649 \\
        G-PSA & 0.5238 \small$\pm$ 1.01 & \textbf{0.8095 \small$\pm$ 0.63} & \textbf{0.3810 \small$\pm$ 0.90} & 0.4762 \small$\pm$ 1.49 & 0.5476 \small$\pm$ 0.56 \\ 
        & 0.0304 & 0.0002 & 0.0881 & 0.0961 & 0.0058 \\
        GPT-PSA & 0.0476 \small$\pm$ 0.90 & -0.0476 \small$\pm$ 1.19 & 0.2381 \small$\pm$ 1.04 & 0.2857 \small$\pm$ 2.01 & 0.1310 \small$\pm$ 0.66 \\ 
        & 0.8249 & 0.8471 & 0.3086 & 0.3786 & 0.6999 \\
        \hline
    \end{tabular}
    \caption{Proposal Quality relative to Rewritten Originals ($\mu \pm \sigma^2$)}
    \label{tab:human_eval_all_vs_original}
\end{table*}

\begin{table*}[bt]
    \centering
    \rowcolors{1}{wtbl}{gtbl}
    \begin{tabular}{l c c c c c}
        \hline
        Model & Appropriateness & Feasibility & Thoroughness & Expected Effectiveness & Average* \\
        \hline\hline
        DeepSeek-V3 vs DS-PSA & \textbf{0.6667 \small$\pm$ 0.70} & \textbf{0.5238 \small$\pm$ 1.30} & \textbf{0.6190 \small$\pm$ 0.81} & \textbf{0.5238 \small$\pm$ 0.92} & \textbf{0.5833 \small$\pm$ 0.48} \\
        & 0.0019 & 0.0530 & 0.0059 & 0.0237 & 0.0190 \\
        Gemini-2.0 vs G-PSA  & -0.0476 \small$\pm$ 0.71 & 0.0476 \small$\pm$ 0.43 & 0.1429 \small$\pm$ 0.50 & -0.2857 \small$\pm$ 0.97 & -0.0357 \small$\pm$ 0.22 \\ 
        & 0.8033 & 0.7477 & 0.3786 & 0.2084 & 0.6100 \\
        GPT-4o vs. GPT-PSA & 0.4286 \small$\pm$ 1.48 & 0.2381 \small$\pm$ 1.61 & 0.4762 \small$\pm$ 1.39 & 0.3333 \small$\pm$ 1.46 & 0.369 \small$\pm$ 0.97 \\ 
        & 0.1312 & 0.4113 & 0.0862 & 0.2317 & 0.4928 \\
        \hline
    \end{tabular}
    \caption{Proposal Quality relative to Base Models ($\mu \pm \sigma^2$)\\ *$p$-values for $H_1: \mu \neq 0$ according to Hotelling's $T^2$-test}
    \label{tab:human_eval_psa_vs_base}
\end{table*}

\begin{table}[hbt]
    \centering
    \begin{tabular}{l c c c}
        \hline
        Model & Base & PSA & Proportion \\
        \hline
        GPT-4o & 34 & 57 & 1.676 \\
        Gemini-2.0 & 31 & 65 & 2.097 \\
        \hline
    \end{tabular}
    \caption{Unique Problem Counts for PSAs vs. Core Models}
    \label{tab:probcounts}
\end{table}

For each PSA and their underlying model, we report the mean difference between metrics from the original proposal and $p$-values in Table \ref{tab:human_eval_all_vs_original}; bold cells indicate the highest performance for that metric. The first row indicates the mean score and variance assigned to each metric for the original, rewritten proposals, and the remaining rows indicate the relative difference. Based on these differences, we observe that the PSAs generate more comparable proposals to the originals, whereas their vanilla counterparts come up a bit short, as seen in the performance of both DeepSeek-V3 and GPT-4o. The margin for DeepSeek-V3 compared to original human proposal is significant, scoring an average of 0.619 lower ($p=0.0017$). 
The exception is Gemini-2.0.
The proposals generated by Gemini-2.0-Flash were scored 0.5833 higher on average ($p=0.0022$) with significant increases in each metric except thoroughness. Gemini-2.0 additionally generated consistently better proposals as a baseline and PSA than both DeepSeek-V3 and GPT-4o. This is likely due to a difference in implicit knowledge as Gemini demonstrated a more nuanced understanding of social issues, allowing for a more detailed problem statement and therefore more cohesive solution. 

To compare the effectiveness of the PSA framework, we report the mean difference between a PSA and its base model in Table \ref{tab:human_eval_psa_vs_base}. There is a clear increase in the proposal quality for both DS-PSA and GPT-PSA while G-PSA generated proposals lower in quality relative to their base models. Among these differences, only DeepSeek-V3 vs. DS-PSA is significant ($p = 0.019$). Following our preliminary experiments with GPT-4o-mini, these results suggest that our framework is most beneficial for models that lack the necessary knowledge for creating an AI4SG proposal. We make several notes about our experimental design in Section 7. 

\subsection{Qualitative Findings \& Considerations}

\begin{figure*}[t!]
\begin{tcolorbox}[boxrule=1pt]
\begin{quote}
\textbf{DeepSeek-V3: AI-Powered Community Engagement Enhancement}

\textbf{Problem Statement}: The New York City Mayor's Public Engagement Unit faces challenges in effectively reaching and engaging with all segments of the city's diverse population, especially those in underserved communities. These communities often have \hl{limited access to digital platforms and face socioeconomic barriers} that hinder their participation in public programs. This lack of engagement leads to underutilization of services and a gap in understanding the needs of these populations. Evidence from recent surveys indicates that over 30\% of residents in underserved areas are unaware of available public services, highlighting the urgency of this issue.

\textbf{Proposed Solution}: To address this, we propose the development of an AI-driven platform that personalizes communication and identifies the most effective channels and times for reaching different demographic groups. The solution leverages natural language processing to \hl{analyze community feedback and sentiment from various sources, including social media, public forums, and direct feedback channels}. Machine learning models will be trained on this data, along with demographic and socioeconomic datasets, to predict the best engagement strategies for different groups. The platform will also use \hl{computer vision to analyze images from community events and public spaces, identifying patterns in participation and engagement}. This approach will enable the Public Engagement Unit to tailor its outreach efforts more effectively, ensuring that all residents, especially those in underserved communities, are informed and engaged with public services. The data sources will include publicly available datasets, social media data, and internal records from the Public Engagement Unit. This solution is designed to be scalable and adaptable to the changing needs of New York City's diverse population.
\end{quote}
\end{tcolorbox}
\caption{DeepSeek-V3 Generated Proposal for New York Mayor's Public Engagement Unit}
\label{nypeu_dsv3}
\end{figure*}

\subsubsection{Models Struggle to Maintain Clear Objectives}

Our baselines occasionally struggle to maintain clear objectives when formulating solutions. This difference, while subtle at times, results in a mismatch between the output of a proposal and its initial motivations. Consider an example from DeepSeek-V3 in Figure \ref{nypeu_dsv3}: Limited access to technology contributes to the under representation of marginalized communities, but the proposed solution intends to leverage social media and other digital platforms to predict ``engagement strategies". It is illogical to propose collecting data when the primary motivating challenge is that the target population has limited access to platforms where such data would be available. Additionally, the use of computer vision for surveilling public spaces could raise ethical concerns from the public as this project would be conducted by a governmental institution. Though not catastrophic errors, it is a detail that should be considered and likely would be by a human expert. We include a DS-PSA-generated proposal for comparison in the Appendix.

\subsubsection{LLMs Lack Diversity during Generation}

As noted in \cite{si2024llmsgeneratenovelresearch}, LLMs struggle with diversity during idea generation. We observe a similar effect during problem scoping where models focus on a similar subset of problems given an organization's domain. To compare the diversity of problems selected by the base LLMs, we sample $5$ proposals per organization for a total of $105$ proposals and select only the unique problems for Gemini-2.0 and GPT-4o. We report the total unique ideas as well as the proportion $x_{PSA}/x_{Base}$in Table \ref{tab:probcounts}, with GPT-PSA and G-PSA selecting roughly 1.6x and 2.1x more unique problems respectively. By leveraging Google's Search API and sampling from a subset of retrieved challenges, we encourage the PSAs to tackle a wider variety of relevant problems.

\section{Future Work}

\subsubsection{Real Problem Scoping is Interactive}

Problem scoping for AI4SG process is generally collaborative, involving experts familiar with both the domain and technical knowledge necessary for enacting an adequate AI solution. In contrast, this work is a proof-of-concept for the application of AI in an AI4SG setting and leverages a single AI agent when generating proposals without any human interaction. Future work may try a Human-AI collaborative approach to overcome the current challenges with AI4SG problem scoping, taking advantage of human reasoning and the expert knowledge of AI models.

\subsubsection{Scoping Social Issues is Subjective}

A recurrent challenge we faced in this work is the subjectivity of problem scoping. Problem importance largely depend on one's knowledge and background, leading to potential disagreements between what proposals are ``better" than one another. As this application of AI is relatively new, AI4SG problem scoping is not well formalized and lacks well-defined criteria and large amounts of data other AI tasks may have. For future work, it may be interesting to test  the subjectivity and preferences of LLMs during problem selection, particularly how it relates to the domain familiarity an LLM has and its agreement with human values. 

\bibliography{aaai25}

\clearpage

\appendix

\begin{figure*}[bt]
\begin{tcolorbox}[boxrule=1pt]
\begin{quote}
\textbf{DS-PSA: Multilingual AI Assistant for Housing Services Access}

\textbf{Problem Statement}: Language barriers significantly hinder non-English speaking New Yorkers from accessing essential housing services, including tenant rights information, eviction prevention, and affordable housing applications. This issue disproportionately affects immigrants and low-income families, who are more likely to face housing instability. The lack of accessible, multilingual resources exacerbates socioeconomic disparities, making it difficult for these communities to navigate the city's housing support systems. Evidence from community feedback and service usage statistics highlights the urgent need for a solution that bridges the language gap in housing services access.

\textbf{Proposed Solution}: We propose the development of a Multilingual AI Assistant that leverages Natural Language Processing (NLP) and Machine Translation (MT) technologies to provide real-time, accurate translations and personalized guidance for housing services. This AI-driven platform will integrate with existing city services, offering support in over 20 languages to ensure inclusivity. The assistant will utilize a combination of pre-trained language models and custom datasets, including housing service FAQs, tenant rights documents, and user interaction logs, to improve translation accuracy and contextual understanding. By employing advanced NLP techniques, the system will understand and respond to user queries in their native language, facilitating easier access to housing resources. The solution addresses the limitations of traditional translation services by continuously learning from user interactions to enhance its performance and adapt to the diverse linguistic needs of New Yorkers.
\end{quote}
\end{tcolorbox}
\caption{DS-PSAfor $O=$ ``New York Mayor's Public Engagement Unit"}
\label{nypeu_deepseek}
\end{figure*}

\section{Prompts}
\label{appendix:prompts}

\subsubsection{System}

You are an artifical intelligence (AI) expert in a group consulting \{organization.name\}.

\subsubsection{Annotator}

[PAPERS]: \{papers\} [TASK]: For each paper, summarize the following information in a paragraph: 1) the problem 2) any methods, models, or techniques used 3) the limitations and restrictions of the mentioned methods/overall work 4) any data used 5) the findings and outcome of the work. Additionally, provide a two numbers between 0 (No confidence, low quality) and 100 (High confidence, high quality) indicating your confidence in the relevance of the paper to your organization and the methods it defines. Be detailed in your summary and provide an explanation for any technical details.

\subsubsection{Background Retrieval}

[ORGANIZATION INFORMATION]: Here are some articles about 
\{organization.name\}: \{articles\}. [TASK]: Summarize \{organization.name\} and its objectives.

\subsubsection{Challenge Retrieval}

[BACKGROUND]: \{background\} [TASK]: What are 5 problems your organization currently faces? Write 5 search queries to find evidence or news for specific problems that affect your organization or the communities you work with. [CONSTRAINT]: Be as specific as possible when searching for challenges. Consider what the common issues are local to your area.

[CHALLENGE SOURCES]: \{challenges\}. From the provided sources, create a list of the critical challenges that \{organization.name\} faces. Include only the unique challenges and their key details and statistics. For each unique challenge, provide two numbers between 0 (Unimportant) and 100 (Very Important) representing your confidence that it the challenge is relevant to the organization and tractable.

\subsubsection{Method Retrieval}

[DOMAIN CHALLENGE]: \{selected\_challenge\}. [TASK]: What machine learning or statistical techniques are appropriate for the provided challenge? Come up with 5 short search queries to find papers that address a challenge with a method you find appropriate.[CONSTRAINT]: Do not include any local information in the query. Search for specific approaches and techniques that are appropriate for the given problem.

\subsubsection{Solution Generation}

[CHALLENGES]: \{selected\_challenge\} [AI METHODS]: \{methods\} [TASK]: For one challenge related to your organization, propose a detailed solution using artificial intelligence. The solution must contain the following: 1) **Title**: a concise project title. 2) **Problem Statement**: This provides a detailed explanation of the problem and why it is important to your organization. It should include: relevant information on those affected by the problem, any notable socioeconomic attributes of the affected group, and evidence for your claims if available. 3) **Proposed Solution**: This outlines the overall approach to the challenge and its relevant AI topics. For each mentioned technical topic, provide a motivation and a high-level explanation. Explain its relevance to the problem in detail; focus on the goal of the approach and how the methods will achieve this goal. The data sources and types (images, records, etc) should be included here, using already available datasets if possible. Avoid solutions that are trivial, purely analytical, or outreach initiatives. The solution should be justifiable and appropriate for any data constraints or types introduced by the problem. [CONSTRAINTS]: This statement must be related to your organization, \{organization.name\}. If there is a topic local to your area or specific to your field, elaborate on it. Assume that the reader will be a non-local, non-expert reviewer. Since you are an expert on this problem, write the proposal in a convincing tone as a group. Consider the limitations of each AI method and the constraints of the challenge. Do not use acronyms or cite papers. Return a concise paragraph for each section as well as a confidence estimate between 0 (No success) and 100 (Definitely successful) in the success of this proposal."

\subsection{Evaluation}
\label{appendix:formats}

\subsubsection{Criteria}

Each metric is rated 1-5 (Likert scale) with a 2-3 sentence rationale. 

**Appropriateness:** How related the proposal and identified problem are to the original domain. Is the identified problem related to the organization's goals? Does this problem necessitate the use of AI methods?
1. The identified problem is irrelevant outside of the organization's domain.
2. The identified problem is relevant but intractable.
3. The identified problem is generic and easily inferrable from the organization's domain.
4. The identified problem is specific but lacks a motivation for the usage of AI.
5. The identified is specific and motivates the usage of AI adequately.

**Thoroughness:** How concrete the proposed solution is. Are any resource requirements specified? Is the solution justified, and does the justification make technical sense? Are the methods tied in with the identified problem (e.g. for resource allocation, what resources are being allocated)?
1. The approach is incomprehensible and does not make any technical sense.
2. The approach is makes technical sense but is not connected to the problem statement.
3. The proposed approach is technically sound but addresses issues outside of the problem statement.
4. The proposed approach is broadly motivated by the target problem but lacks detail.
5. The proposed approach is connected to the target problem with specific examples and motivations.

**Feasibility:** Is the identified problem reasonably tractable? How much progress can be made within the first few weeks? What are the metrics for success as the project develops? Are the requirements of the solution reasonable for the organization?
1. The identified problem is intractable.
2. The identified problem is tractable but unreasonable for the scope of the organization.
3. The proposed approach will largely depend on external factors, like other organizations or human interventions. 
4. The proposed approach could solve the problem with additional resources.
5. The proposed approach can sufficiently solve the problem with existing resources.

**Expected Effectiveness:** The impact on the identified problem given a successful implementation of the proposed solution. What will change, and how much will it change? Is this project sustainable? Will this solution continue to be useful for the target demographics within the next decade?
1. The proposed approach will have no influence on the identified problem.
2. The resources required for the proposed approach exceeds the potential impact.
3. The proposed approach roughly matches the resource requirements but may be beneficial in the long run.
4. The proposed approach is resource-efficient and will have an impact on the target demographics but has a large risk for bias.
5. The proposed approach is resource-efficient and will have a significant impact on the target demographics and necessitates few ethical considerations.

\subsection{Mean Differences According to AI Evaluations}
\label{appendix:ai_evaluations}

\begin{table*}[bt]
    \centering
    \rowcolors{1}{wtbl}{gtbl}
    \begin{tabular}{l c c c c c}
        \hline
        Model & Appropriateness & Feasibility & Thoroughness & Expected Effectiveness & Average* \\
        \hline\hline
        Original & 4.9683 $\pm$ 0.00 & 4.0317 $\pm$ 0.00 & 3.7249 $\pm$ 0.07 & 4.1746 $\pm$ 0.08 & 4.2249 $\pm$ 0.02 \\
        \hline
        DeepSeek-V3 & -0.0106 $\pm$ 0.03 & -0.0106 $\pm$ 0.01 & 0.0265 $\pm$ 0.14 & -0.0582 $\pm$ 0.16 & -0.0132 $\pm$ 0.05 \\
        & 0.7711 & 0.5764 & 0.7509 & 0.5189 & 0.7900 (0.6326) \\
        Gemini-2.0-Flash & 0.0159 $\pm$ 0.01 & 0.1481 $\pm$ 0.04 & 0.0635 $\pm$ 0.09 & -0.0794 $\pm$ 0.07 & 0.037 $\pm$ 0.02 \\
        & 0.3293 & 0.0032 & 0.3613 & 0.1861 & 0.2038 (0.0072) \\
        GPT-4o & -0.0106 $\pm$ 0.03 & -0.0106 $\pm$ 0.01 & 0.0265 $\pm$ 0.14 & -0.0582 $\pm$ 0.16 & -0.0132 $\pm$ 0.05 \\
        & 0.7711 & 0.5764 & 0.7509 & 0.5189 & 0.7900 (0.6326) \\
        \hline 
        DS-PSA & -0.0 $\pm$ 0.01 & 0.0106 $\pm$ 0.01 & -0.0476 $\pm$ 0.19 & -0.0635 $\pm$ 0.08 & -0.0251 $\pm$ 0.03 \\
        & 1.000 & 0.6657 & 0.6272 & 0.3151 & 0.5261 (0.8830) \\
        G-PSA & 0.0159 $\pm$ 0.00 & 0.1693 $\pm$ 0.07 & 0.0529 $\pm$ 0.18 & -0.0582 $\pm$ 0.13 & 0.0450 $\pm$ 0.04 \\
        & 0.1861 & 0.0077 & 0.5802 & 0.4749 & 0.3474 (0.002) \\
        GPT-PSA & -0.0212 $\pm$ 0.02 & -0.0212 $\pm$ 0.01 & 0.0423 $\pm$ 0.14 & -0.0688 $\pm$ 0.11 & -0.0172 $\pm$ 0.03 \\
        & 0.5501 & 0.2579 & 0.6161 & 0.372 & 0.6813 (0.1681) \\
        \hline
    \end{tabular}
    \caption{(AI) Differences to Baseline ($\mu \pm \sigma^2$) \\ *$p$-values for $H_A \mu \neq 0$ according to Paired $t$-test ($T^2$-test)}
    \label{tab:my_label}
\end{table*}

\end{document}